\documentclass[10pt,twocolumn,letterpaper]{article}

\usepackage{cvpr}
\usepackage{times}
\usepackage{epsfig}
\usepackage{graphicx}
\usepackage{amsmath}
\usepackage{amssymb}
\usepackage[breaklinks=true,bookmarks=false]{hyperref}

\cvprfinalcopy % *** Uncomment this line for the final submission

\def\cvprPaperID{****} % *** Enter the CVPR Paper ID here

\begin{document}

%%%%%%%%% TITLE
\title{You Look Twice: GaterNet for Dynamic Filter Selection in CNNs}

% four groups
% \author{
% Zhourong Chen\thanks{Student of The Hong Kong University of Science and 
% Technology (HKUST). Work is done when interning at Google.}\\
% Google Research, Mountain View\\
% HKUST, Hong Kong\\
% {\tt\small zchenbb@cse.ust.hk}
% \and
% Yang Li\\
% Google Research, Mountain View\\
% {\tt\small liyang@google.com}
% \and
% Samy Bengio\\
% Google Brain, Mountain View\\
% {\tt\small bengio@google.com}
% \and
% Si Si\\
% Google Research, Mountain View\\
% {\tt\small sisidaisy@google.com}
% }

% Two groups:
% \author{
% Zhourong Chen\thanks{Student of The Hong Kong University of Science and 
% Technology. Work is done when interning at Google.}\\
% Google Research, Mountain View\\
% The Hong Kong University of Science and Technology, Hong Kong\\
% {\tt\small zchenbb@cse.ust.hk}
% \and
% Yang Li, Samy Bengio, Si Si\\
% Google Research, Mountain View\\
% {\tt\small \{liyang, bengio, sisidaisy\}@google.com}
% }

% One group:
\author{
Zhourong Chen$^{1,2}$\thanks{Student of The Hong Kong University of Science and 
Technology. Work is done when interning at Google.}, \quad Yang Li$^1$, \quad Samy Bengio$^1$, \quad Si Si$^1$\\
$^1$Google Research, Mountain View\\
$^2$The Hong Kong University of Science and Technology, Hong Kong\\
{\tt\small zchenbb@cse.ust.hk, \{liyang, bengio, sisidaisy\}@google.com}
}

\maketitle
%\thispagestyle{empty}

%%%%%%%%% ABSTRACT
\begin{abstract}
The concept of conditional computation for deep nets has been proposed previously to improve model performance by selectively using only parts of the model conditioned on the sample it is processing.
In this paper, we investigate input-dependent dynamic filter selection in deep convolutional neural networks (CNNs).
The problem is interesting because the idea of forcing different parts of the model to learn
from different types of samples may help us acquire better filters in CNNs, improve the model
generalization performance and potentially increase the interpretability of model behavior.
We propose a novel yet simple framework called GaterNet, which 
involves a backbone and a gater network. The backbone network is a regular CNN that performs the major
computation needed for making a prediction, while a global gater network is introduced to generate 
binary gates for selectively activating filters in the backbone network based on each input.
Extensive experiments on CIFAR and ImageNet datasets
show that our models consistently
outperform the original models with a large margin. On CIFAR-10, our model also
improves upon state-of-the-art results.
\end{abstract}

%%%%%%%%% BODY TEXT
\section{Introduction}
It is widely recognized in neural science that distinct parts of the brain are highly specialized for different types of 
tasks~\cite{kandel1991principles}.
It results in not only the high efficiency in handling a response but also the surprising effectiveness of the brain in learning new events. In machine learning, {\it conditional computation}~\cite{bengiolooking} has been proposed to have 
a similar mechanism in deep learning models. 
For each specific sample, the basic idea of conditional computation is to only involve a small portion 
of the model in prediction. It also means that only a small fraction of parameters needs to be updated at 
each back-propagation step, which is desirable for training a large model.

\begin{figure}[t]
\begin{center}
   \includegraphics[width=\linewidth]{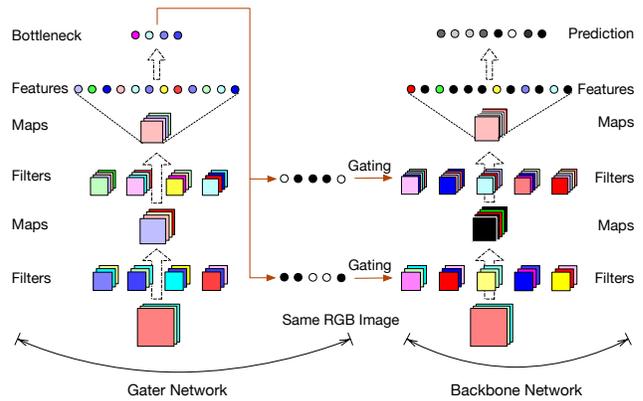}
\end{center}
   \caption{Model Architecture (better viewed in color). 
   The gater extracts features and generates sparse 
   binary gates for selecting
   filters in the backbone network in an input-dependent manner.
   }
\label{figs.model}
\vspace{-5mm}
\end{figure}

One line of work that is closely related to conditional computation is {\it Mixture of Experts} 
(MoE)~\cite{jacobs1991adaptive}, where multiple sub-networks are combined via an ensemble using weights determined by a gating module. Particularly, several recent works~\cite{outrageously,Mullapudi_2018_CVPR} propose to ensemble a small subset of dynamically selected experts in the model for each input.
By doing so, these models are able to reduce
computation cost while achieving similar or even better results than baseline models. Note that both the expert architectures and the number of experts in these works are pre-defined and fixed.
Another line of 
works that resemble conditional computation 
focus on {\it dynamic network configuration} ~\cite{veitadaptive,ba2013adaptive,denoyer2014deep,bengio2015conditional,bengio2013estimating}. 
There are no explicitly defined experts in these methods.
Rather, they dynamically select the units, layers, or other components in the main model for each input.
In these works, one small sub-module is usually added to each position to be configured in the model. That is, each sub-module added is making decisions locally specific to the components it is configuring.

In this paper, we propose a novel framework called {\it GaterNet} for input-dependent dynamic filter selection 
in convolutional neural networks (CNNs), as shown in Figure~\ref{figs.model}.
We introduce a dedicated sub-network called {\it gater}, which extracts features from input and
generates the binary gates needed for controlling filters all at once based on the features.
The gating vector is then used to select the filters in the {\it backbone}\footnote{The term 
{\it backbone} is also used in the literature of object detection and TSE-Net~\cite{tsenet}.}
network (the main model in our framework), and only the selected filters in
the backbone network participate in the prediction and learning.
We used a discretization technique called 
{\it Improved SemHash}~\cite{kaiser2018discrete} to enable differentiable training of
input-dependent binary gates such that the backbone and the gater network can be trained jointly
via back-propagation. 

Compared to previous works on dynamic 
network configuration, we use a dedicated sub-network (the gater) for making global decisions on which filters in the backbone network should be used. The decision on each gate (for each filter) is made based on a shared global view of the current input. 
We argue that such a global gating unit can make more holistic decisions about how to optimally use the filters in the network than local configuration employed by previous work.
Note that in~\cite{Mullapudi_2018_CVPR}, a module of the network is used to generate all the gates, which at a glance is similar to our gater. However, there are two important differences. 
Firstly,~\cite{Mullapudi_2018_CVPR} is not based on an end-to-end approach. It requires a pre-processing step to cluster classes of samples and assign each cluster to a sub-branch of the network to handle. The assignments provides explicit supervision for training the gating module. Secondly, as mentioned above, the sub-branch architectures and the number
of branches are both manually defined and fixed throughout training in~\cite{Mullapudi_2018_CVPR}. In contrast, in our framework, each sample uses a dynamically determined sub-branch depending on the filters being selected. As a result, our method potentially allows a combinatorial number of choices of sub-branches or experts given the number of filters to be controlled, which is more amenable for capturing complex distribution manifested in the data.

Our experiments on CIFAR~\cite{krizhevsky2009learning} 
and ImageNet~\cite{ILSVRC15} classification datasets
show that the gater in GaterNet is able to learn effective gating strategies for selecting
proper filters. It consistently improves the original model
with a significant margin.
On CIFAR-10, our method gives better classification results 
than state-of-the-art models with only 1.2\% additional parameters. 
Our contributions are summarized as follows:
\begin{itemize}
    \vspace{-3mm}
    \item We propose a new framework for dynamic filter selection in CNNs. The core of the idea is to introduce a dedicated gater network to take a glimpse of the input, and then generate input-dependent binary gates to select filters in the backbone network for processing the input. By using Improved SemHash, the gater network can be trained jointly with the backbone in an end-to-end fashion through back-propagation.
    \vspace{-3mm}
    \item We conduct extensive experiments on GaterNet, which show that it consistently improves the generalization performance of deep CNNs without significantly increasing the model complexity. In particular, our models achieve better results than several state-of-the-art models on the CIFAR-10 dataset by only introducing a small fraction of parameters.
    \vspace{-3.5mm}
    \item We perform an in-depth analysis about the model behavior for GaterNet, which reveals that GaterNet learns effective gating strategies by being relatively
    deterministic on the choice of filters to use in shallow layers but using more input-dependent filters in the deep layers. 
\end{itemize}

%------------------------------------------------------------------------
\section{Related Work}
The concept of conditional computation is first discussed by Bengio in~\cite{bengiolooking}.
Early works on conditional computation focus on how to
select model components on the fly. 
Bengio et al. have studied four approaches for learning stochastic neurons in fully-connected neural networks for conditional selection in~\cite{bengio2013estimating}.
On the other hand, Davis and Arel have used low-rank approximations to predict the sparse activations of neurons at each layer \cite{davis2013low}. 
Bengio et al. have also tested reinforcement learning to optimize conditional
computation policies~\cite{bengio2015conditional} .

More recently, Shazeer et al. have investigated the combination of conditional computation with Mixture of Experts on language modeling and machine translation tasks \cite{outrageously}. At each time step in the sequence model, they dynamically select a small subset of experts to process the input. 
Their models significantly outperformed state-of-the-art models with a low computation cost. In the same vein, Mullapudi et al. have proposed HydraNets that uses multiple branches of networks for extracting features \cite{Mullapudi_2018_CVPR}. In this work, a gating module
is introduced to generate decisions on selecting branches for each specific input.
This method requires a pre-processing step of clustering the ground-truth classes to force each branch to learn features for a specific cluster of classes as discussed in the introduction.

Dynamic network configuration is another type of conditional computation that has been studied previously. In this line of works, no parallel experts are explicitly defined.
Instead, they dynamically configure a single network by selectively activating model components such as units and layers for each input.
Adaptive Dropout is proposed by Ba and Frey to dynamically learn a dropout rate for 
each unit and each input~\cite{ba2013adaptive}. Denoyer and Ludovic have
proposed a tree structure neural network called Deep Sequential Neural Network \cite{denoyer2014deep}. A path from the root to a leaf node in the tree represents
a computation sequence for the input, which is also dynamically determined for each input.
Recently, Veit and Belongie \cite{veitadaptive} have proposed to skip layers in ResNet~\cite{he2016deep} 
in an input-dependent manner. The resulting model is performing better and also more
robust to adversarial attack than the original ResNet, which also leads to reduced computation cost.

Previous works have also investigated methods that dynamically re-scale or calibrate the different components in a model. The fundamental difference between these methods and dynamic network 
configuration is that they generate a real-valued vector for each input, instead of a
binary gate vector for selecting network components.
SE-Net proposed by Hu et al.~\cite{hu2018senet} re-scales the channels in feature maps on the fly
and achieves state-of-the-art results on ImageNet classification dataset.
Stollenga et al.~\cite{stollenga2014deep} have also proposed to go through the main model for multiple passes.
The features resulting from each pass (except the last) are used to 
generate a real-valued vector for re-scaling the channels in the next pass. In contrast to these works, our gater network generates binary decisions to dynamically turn on or off filters depending on each input.

%------------------------------------------------------------------------
\section{GaterNet}
Our model contains two convolutional neural sub-networks,
namely the {\it backbone network} and the {\it gater network} as illustrated in Figure~\ref{figs.model}. Given an input, the gater network decides the set of filters in the backbone network for use while the backbone network does the actual prediction. The two sub-networks are trained in an end-to-end manner via back-propagation.
\subsection{Backbone}
The backbone network is the main module of our model, which extracts features from input
and makes the final prediction.
Any existing CNN architectures such as ResNet~\cite{he2016deep}, 
Inception~\cite{Szegedy2017Inceptionv4IA} and DenseNet~\cite{huang2017densely} 
can be readily used as the backbone network in our GaterNet.

Let us first consider a standalone backbone CNN without the gater network.
Given an input image $x$, the output of the $l$-th convolutional layer is a 3-D 
feature map $O^l(x)$.
In a conventional CNN, $O^l(x)$ is computed as:
\begin{equation}\label{eq.fea_map}
O^l_i(x) = \phi(F^l_i * I^l(x)),
\end{equation}
where $O^l_i(x)$ is the $i$-th channel of feature map $O^l(x)$, $F^l_i$ is the $i$-th
3-D filter, 
$I^l(x)$ is the 3-D input feature map to the $l$-th layer, % I not necessarily the previous layer
$\phi$ denotes the element-wise nonlinear activation function,
and $*$ denotes convolution. 
In general cases without the gater network, all the filters $F^l_i$ in the current layer
are applied to $I^l(x)$, 
resulting in a {\it dense} feature map $O^l(x)$.
The loss for training such a CNN for classification is
$
L = -\log P(y|x, \theta)
$
for a single input image, where $y$ is the ground-truth label and $\theta$ denotes the 
model parameters.

\subsection{Gater}
In contrast to the backbone, the gater network is an 
assistant of the backbone and does not learn any features directly used in 
the prediction. Instead, the gater network 
processes the input to generate an input-dependent {\it gating mask---a binary vector}. 
The vector is then used to dynamically 
select a particular subset of filters in the backbone network for the current input.
Specifically, the gater network learns a function as below:
\begin{equation}\label{eq.gater}
G(x) = D(E(x)).
\end{equation}
Here, $E$ is an image feature extractor defined as $E: x \rightarrow f, x \in 
\mathbb{R}^{h'\times w'\times c'}, f \in \mathbb{R}^{h}$, with $h', w', c'$ being the 
height, width and channel number of an input image respectively, and $h$ being the 
number of features extracted. $D$ is a function defined as $D: f \rightarrow g, f \in 
\mathbb{R}^{h}, g \in \{0, 1\}^{c}$, where $c$ is the total number of filters in the 
backbone network. More details about function $E$ and $D$ will be discussed in 
Section~\ref{sec.fea_extract} and Section~\ref{sec.gates} respectively.

From the above definition we can see that, the gater network learns a function which maps
input $x$ to a binary gating vector $g$. With the help of $g$,
we reformulate the computation of feature map $O^l(x)$ in~Equation (\ref{eq.fea_map}) as 
below:
\begin{equation}\label{eq.selected_fea_map}
O^l_i(x) =  \begin{cases}
                {\bf 0},                & \text{if } g^l_i = 0\\
                \phi(F^l_i * I^l(x)),   & \text{if } g^l_i = 1
            \end{cases}
\end{equation}
Here $g^l_i$ is the entry in $g$ corresponding to the $i$-th filter at layer $l$,
and ${\bf 0}$ is a 2-D feature map with all its elements being 0. 
That is, the $i$-th filter will be applied to $I^l(x)$ to extract features only 
when $g^l_i=1$. If $g^l_i=0$, the $i$-th filter is skipped and ${\bf 0}$ is used as the output instead. When $g^l$ is a sparse binary vector, a large subset of filters will be 
skipped, resulting in a sparse feature map.
In this paper, we implement the computation in Equation (\ref{eq.selected_fea_map}) 
by masking the output channels using the binary gates:
\begin{equation}\label{eq.masked_fea_map}
O^l_i(x) = \phi(F^l_i * I^l(x)) \cdot g^l_i
\end{equation}

In the following subsections, we will introduce how we design the functions $E$ and $D$ in
Equation~(\ref{eq.gater}) and how we enable end-to-end training through the binary gates.

\vspace{-4mm}
\subsubsection{Feature Extractor}\label{sec.fea_extract}
\vspace{-2mm}
Essentially, the function $E(x)$ in Equation (\ref{eq.gater}) is a feature extractor which 
takes an image $x$ as input and outputs a feature vector $f$. Similar to the backbone network,
any existing CNN architectures can be used here to learn the function $E(x)$. 
There are two main differences compared with the backbone network: (1) The output layer of the CNN architecture is removed 
such that it outputs features for use in the next step.
(2) A gater CNN does not necessarily need to be as complicated as the one for the backbone. 
One reason is that the gater CNN is supposed to obtain a brief view of the input.
Having an over-complicated gater network may encounter various difficulties in computation cost and 
optimization. Another reason is to avoid the gater network accidentally taking over the task that is intended for the backbone network.

\subsubsection{Features to Binary Gates}\label{sec.gates}
\vspace{-2mm}
\paragraph{Fully-Connected Layers with Bottleneck}
As defined in Equation~(\ref{eq.gater}), the function $D(f)$ needs to map the vector $f$ 
of size $h$ to a binary vector $g$ of size $c$. We first consider using 
fully-connected layers to map $f$ to a {\it real-valued} vector $g'$ of size $c$. 
If we use one single layer to project the vector, the projection matrix would be of size
$h\times c$.
This can be very large when $h$ is thousands and $c$ is tens of thousands. 
To reduce the number of parameters in this projection, we use two
fully-connected layers to fulfill the projection. The first layer projects $f$ to a bottleneck
of size $b$, followed by the second layer mapping the bottleneck to $g'$. 
In this way, the total number of parameters becomes $(h+c)\times b$. 
This can be significantly smaller than $h \times c$ when $b$ is much
smaller than $h$ and $c$. We ignore bias parameters here for simplicity.

In summary, the real-valued vector $g'$ is computed as:
\begin{align*}
    f' &= FC_1(f) \\
    g' &= FC_2(ReLU(BatchNorm(f')))
\end{align*}
where $FC_1$ and $FC_2$ denotes the two linear projections, 
ReLU denotes the non-linear activation function in~\cite{nair2010rectified}, 
and BatchNorm means batch normalization~\cite{batchnorm}.

\vspace{-4mm}
\paragraph{Improved SemHash}\label{para.semhash}
So far, one important question still remains unanswered: how to generate binary gates $g$ 
from $g'$ such that we can back-propagate the error through the
discrete gates to the gater?
In this paper, we adopt a method 
called {\it Improved SemHash}~\cite{kaiser2018discrete,pmlr-v80-kaiser18a}.

During training, we first draw noise from a $c$-dimentional Gaussian distribution with mean 0 and 
standard deviation 1. The noise $\epsilon$ is added to $g'$ to get a noisy version of the
vector: $g'_{\epsilon} = g' + \epsilon$. 
Two vectors are then computed from $g'_{\epsilon}$: 
$$
g_{\alpha} = \sigma '(g'_{\epsilon}) \text{ and } g_{\beta} = {\bf 1}(g'_{\epsilon} > 0) 
$$
where $\sigma'$ is the saturating sigmoid function~\cite{kaiser2015neural,kaiser2016can}:
$$
\sigma'(x) = max(0, min(1, 1.2\sigma(x)-0.1))
$$
with $\sigma$ being the sigmoid function. 
Here, $g_{\alpha}$ is a real-valued gate vector with all the entries falling in the 
interval $[0.0, 1.0]$, while $g_{\beta}$ is a binary vector.
We can see that, $g_{\beta}$ has the desirable binary property that we want to use in our 
model, but the gradient of $g_{\beta}$ w.r.t $g'_{\epsilon}$ is not defined.
On the other hand, the gradient of $g_{\alpha}$ w.r.t $g'_{\epsilon}$ is well defined, but 
$g_{\alpha}$ is not a binary vector.
In forward propagation, we randomly use $g = g_{\alpha}$ for half of the training samples
and use $g = g_{\beta}$ for the rest of the samples.
When $g_{\beta}$ is used,
we follow the solution in~\cite{kaiser2018discrete,pmlr-v80-kaiser18a} and
define the gradient of $g_{\beta}$ w.r.t $g'_{\epsilon}$ to be the same as the gradient of 
$g_{\alpha}$ w.r.t $g'_{\epsilon}$ in the backward propagation.

The above procedure is designed for the sake of easy training.
Evaluation and inference are different to the training phase in two aspects.
Firstly, we skip the step of drawing noise and always set $\epsilon=0$.
Secondly, we always use the discrete gates $g = g_{\beta}$ in forward propagation. 
That is, the gate vector is always binarized in evaluation and inference phase.
The interested readers are referred to~\cite{kaiser2018discrete,pmlr-v80-kaiser18a} for more intuition
behind Improved SemHash.

We use binary gates other than attention~\cite{xu2015show} or other real-valued gates for two reasons. Firstly, binary gates can completely deactivate some filters for each input, and hence those filters will not be influenced by the irrelevant inputs. This may lead to training better filters than real-valued gates. Secondly, discrete gates open the opportunity for model compression in the future.

\vspace{-4mm}
\paragraph{Sparse Gates}
To encourage the gates $g$ to be sparse, we introduce a $L_1$
regularization term into the training loss:
$$
L = -\log P(y|x, \theta) + \lambda \frac{\|G(x)\|_1}{c}
$$
where $\lambda$ is the weight for the regularization term and $c$ is the size of $g$.
Note that the backbone network receives no gradients from the second term, while the gater network
receives gradients from both the two terms.

\subsection{Pre-training}
While our model architecture is straightforward, there are several empirical challenges to train it well. First, it is difficult to learn these gates, which are discrete latent representations. Although Improved SemHash has been shown to work well in several previous works, it is 
unclear whether the approximation of gradients mentioned above is a good solution in our model. 
Second, the introduction of gater network into the model has essentially changed the
optimization space. The current parameter initialization and optimization technique may not be suitable for our model. 
We leave the exploration of better binarization, initialization and optimization techniques 
to our future works. In this paper, we always initialize our backbone network and gater network
from networks pre-traiend on the same task, and empirically find it works well with
a range of models.

%------------------------------------------------------------------------
\section{Experiments}
We first conduct preliminary experiments on CIFAR~\cite{krizhevsky2009learning}
with ResNet~\cite{he2016deep, he2016identity}, which gives us a good understanding
about the performance improvements our method can achieve and also the gating
strategies that our gater is learning.
Then we apply our method to state-of-the-art models on CIFAR-10 and show that we
consistently outperform these models.
Lastly, we move on to a large-scale classification dataset, ImageNet 2012~\cite{ILSVRC15},
and show that our method 
significantly improves the performance of large models, such as ResNet and
Inception-v4~\cite{Szegedy2017Inceptionv4IA}, as well.
\subsection{Datasets}
CIFAR-10 and CIFAR-100 contain natural images belonging to 10 and 100 classes 
respectively. There are 50,000 training and 10,000 test images. We randomly hold out 
5,000 training images as a validation set. All the final results reported on test images are using models
trained on the complete training set. The raw images are with $32\times32$ pixels and we
normalize them using the channel means and standard deviations. 
Standard data augmentation by 
random cropping and mirroring are applied to the training set.
ImageNet 2012 classification dataset contains 1.28 million training images and 50,000 
validation images of 1,000 classes. We use the same data augmentation method as the 
original papers of the baseline models in Table~\ref{tab.err-imagenet}. The images are
of $224\times224$ and $299\times299$ in ResNet and Inception-v4 respectively.
\subsection{Cifar-10 and CIFAR-100}
\vspace{-1mm}
\subsubsection{Preliminary Experiments with ResNet} 
\vspace{-2mm}
We first validate the effectiveness of our method using ResNet as the backbone
network on CIFAR-10 and CIFAR-100 datasets. We consider a shallow version, 
ResNet-20, and two deep versions, ResNet-56 and ResNet-164\footnote{Our ResNet-164 is slightly 
different to the one in~\cite{he2016identity}. The number of filters in the first group of residual 
units are 16, 4, 16 respectively.} 
to gain a better understanding on how our gating strategy can help models with varying capacities. All our gated models employ ResNet-20 as the gater network.
Table~\ref{tab.resnet_cifar} shows the comparison with baseline models on the test set. 
{\it ResNet-Wider} is the ResNet with additional filters at each layer such that it
contains roughly the same number of parameters as our model. {\it ResNet-SE} is the 
ResNet with squeeze-and-excitation block~\cite{hu2018senet}.  
The {\it Gated Filters} column shows the number of filters under consideration in our models.
\vspace{-4mm}
\paragraph{Classification Results}
From the table we can see that, our model consistently outperforms the
original ResNet with a significant margin. On CIFAR100, the error rate of
ResNet-164 is reduced by 1.83\%. 

It is also evident that, our model is performing better than ResNet-SE in all cases. 
Note that our gater network is generating {\it binary gates} for the backbone network 
channels, while ResNet-SE is re-scaling the channels.
It is interesting that, although our method is causing more information 
loss in the forward pass of backbone network due to the sparse 
discrete gates, our model 
still achieves better generalization performance
than ResNet and ResNet-SE. This to some extent validates
our assumption that only a subset of filters are needed for the backbone to
process an input sample.

\vspace{-4mm}
\paragraph{Ablation Analysis on Model Size}
In all cases, ResNet-Wider is better than the original ResNet as well.
ResNet-20-Wider is even the best among all the shallow models. We hypothesize 
that ResNet-20 is suffering from underfitting due to its small amount of 
filters and hence adding additional filters significantly improves the model.
On the other hand, although ResNet-20-Gated has a similar number of parameters
as ResNet-20-Wider, a significant portion (about a half) of its parameters belongs to the gater network, rather than directly participating in prediction, and ResNet-20-Gated still performed on par with ResNet-20-Wider.

The backbone network in ResNet-20-Gated suffers from underfitting 
due to the lack of effective filters. The comparison among the deep models 
validates our hypothesis. ResNet-50 and ResNet-164 contain many more filters than 
ResNet-20, and adding filters to them shows only a minor improvement (see ResNet-50-Wider 
and ResNet-164-Wider). In these cases, our models show a significant improvement over the 
wider models and are the best among all the deep models on both datasets. The comparison
with ResNet-Wider shows that the effectiveness of our model is not solely due to
the increase of parameter number, but mainly due to our new gating mechanism.

\begin{table*}[t]
\caption{Classification error rates on the CIFAR-10 and CIFAR-100 test set. All the methods are with data augmentation. ResNet-Wider is the ResNet with additional filters
at each layer such that it contains roughly the same number of parameters as our 
model. ResNet-SE is the ResNet with squeeze-and-excitation blocks. All the ResNet-Gated
models are using ResNet-20 as the gater network. The Gated Filters column shows the number of
filters subject to gating in our model. 
All the baseline results are from our reimplementation.}
\vspace{-2mm}
\label{tab.resnet_cifar}
\begin{center}
\begin{tabular}{l|r|ll|ll}
& & \multicolumn{2}{c|}{\bf Cifar10}  &\multicolumn{2}{c}{\bf Cifar100} \\ 
& \multicolumn{1}{c|}{\bf Gated Filters} &\multicolumn{1}{c}{\bf Param} & \multicolumn{1}{c|}{\bf Error Rates \%} & \multicolumn{1}{c}{\bf Param} & \multicolumn{1}{c}{\bf Error Rates \%} \\ \hline 
{\bf ResNet-20} \cite{he2016deep}        &- & 0.27M & 8.06 & 0.28M & 32.39 \\
{\bf ResNet-20-Wider}                    &- & 0.56M & {\bf 6.85} & 0.57M & {\bf 30.08} \\
{\bf ResNet-20-SE} \cite{hu2018senet}    &- & 0.28M & 7.81 & 0.29M & 31.22 \\
{\bf ResNet-20-Gated (Ours)}             &336 & 0.55M & {\bf 6.88} ($\downarrow$1.18) & 0.60M & 30.79 ($\downarrow$1.60) \\ \hline
{\bf ResNet-56} \cite{he2016deep}        &- & 0.86M & 6.74 & 0.86M & 28.87 \\
{\bf ResNet-56-Wider}                    &- & 1.08M & 6.72 & 1.09M & 28.39 \\
{\bf ResNet-56-SE} \cite{hu2018senet}    &- & 0.88M & 6.27 & 0.89M & 28.00 \\
{\bf ResNet-56-Gated (Ours)}             &1,008 & 1.14M & {\bf 5.72} ($\downarrow$1.02) & 1.14M & {\bf 27.71} ($\downarrow$1.16) \\ \hline
{\bf ResNet-164} \cite{he2016identity}   &- & 1.62M & 5.61 & 1.64M & 25.39 \\
{\bf ResNet-164-Wider}                   &- & 2.04M & 5.57 & 2.07M & 24.80 \\
{\bf ResNet-164-SE} \cite{hu2018senet}   &- & 2.00M & 5.51 & 2.02M & 23.83 \\
{\bf ResNet-164-Gated (Ours)}            &7,200 & 1.96M & {\bf 4.80} ($\downarrow$0.81) & 1.98M & {\bf 23.56} ($\downarrow$1.83) 
\end{tabular}
\end{center}
\vspace{-7mm}
\end{table*}

\vspace{-4mm}
\paragraph{Complexity}
%FLOPs
It appears to be an issue at a glance if a comprehensive gater network is needed to assist a backbone network, as it may greatly increase the number of parameters. 
However, our experiments show that the gater network does not need to be complex, and as a matter of fact, it can be much smaller than the backbone network (see Table~\ref{tab.resnet_cifar}). 
Although the number of
filters (in the backbone network) under consideration varies from 336 to 7200, the results show that 
a simple gater network such as ResNet-20
is powerful enough to learn input-dependent gates for the three models that have a wide range of 
model capacity. As such, when the backbone network is large (where our method shows more significant 
improvements over baselines), the parameter overhead introduced by the gater network becomes small. For example, ResNet-164-Gated 
has only 20\% more parameters than ResNet-164. In contrast, in other more complicated backbone networks
such as DenseNet and Shake-Shake, this overhead is reduced
to 1.2\% as shown in Table~\ref{tab.sota_cifar}. Consequently, the complexity and the number of additional parameters that our method brings to an existing model is relatively small, especially to large models.

\vspace{-4mm}
\paragraph{Gate Distribution}
One question that would naturally occur is how the distribution of the learned gates looks like. Firstly, it is possible that the gater network is just randomly pruning the backbone network and 
introducing regularization effects similar to dropout into the backbone. It is
also possible that the gates are always the same for different samples.
Secondly, the generated gates may give us good insights into the importance of
filters at different layers.

\begin{figure}[t]
\begin{center}
  \includegraphics[width=\linewidth]{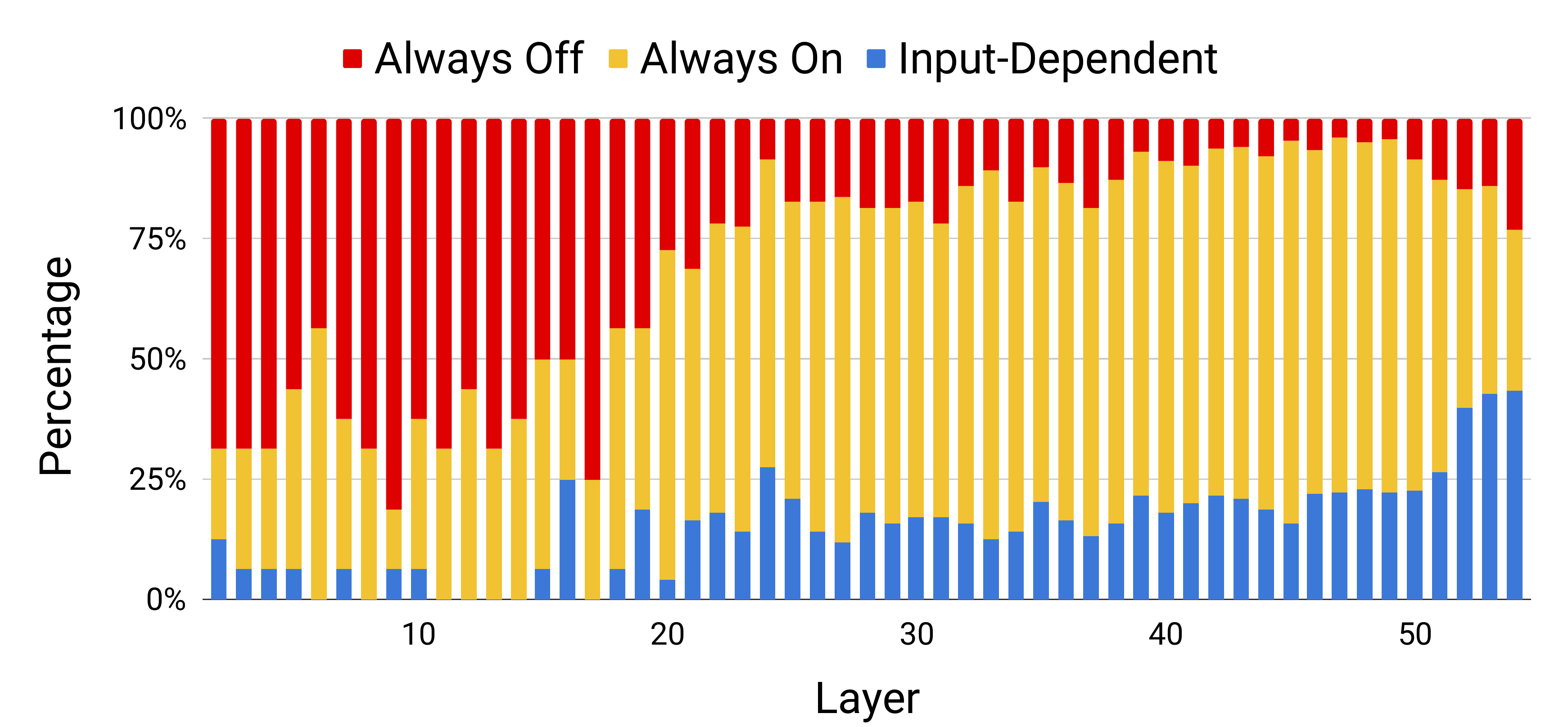}
\end{center}
\vspace{-3mm}
  \caption{The distribution of Gates in each layer for ResNet-164-Gated on Cifar-10 Test Set. There are totally 54 residual units.}
\label{figs.gate_dist}
\vspace{-4mm}
\end{figure}

To answer these questions, we analyzed the gates generated by the gater network in ResNet-164-Gated. We first
conduct forward propagation in the gater network on CIFAR-10 test set and 
collect gates for all the test samples. As expected, three types of gates emerge: 
gates that are always on for all the samples, gates that are always off, and gates 
that can be on or off conditioned on the input, i.e., input-dependent gates.
We show the percentage of the three types of gates at different depth in Figure~\ref{figs.gate_dist}. 
We can 
see that, a large subset (up to 68.75\%) of the gates are always off at the shallow 
residual blocks. As the backbone network goes deeper, the proportion of always-on and 
input-dependent gates increases gradually. In the last two residual blocks, 
input-dependent gates
become the largest subset of gates with percentages of around 45\%. 
The phenomenon is consistent with the common belief that shallow layers are usually 
extracting low-level 
features which are essential for all kinds of samples, while deep layers are
extracting high-level features which are very sample-specific.

\begin{figure}[t]
\begin{center}
   \includegraphics[width=0.8\linewidth]{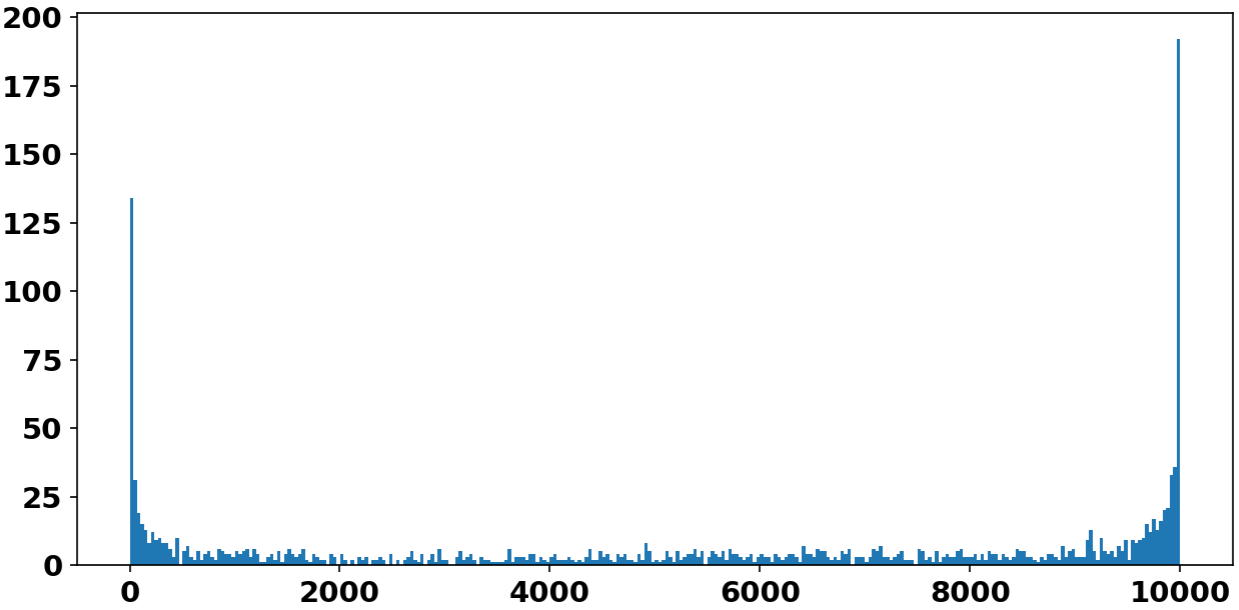}
\end{center}
\vspace{-3mm}
   \caption{Distribution: X-axis is the number of times an input-dependent gate is on, 
   while Y-axis is the number of gates.}
\label{figs.gate_vs_sample}
\end{figure}

Although the above figures show that the gater network is learning input-dependent 
gates, it does not show how often that these gates are on/off. For example, a gate that is on for only one test sample but off for the rest would also appear input-dependent. To investigate this further, we collect all the input-dependent gates and 
plot the distribution of number of times that they are on in 
Figure~\ref{figs.gate_vs_sample}. 
There are totally 1567 input-dependent gates out of the total number of 7200 gates for the backbone network.
While many of these gates remain in one state---either on or off---in most of the time, there are 1,124 gates that switch on and off more frequently---they are activated for 100 $\sim$ 9900 samples out of the 10,000 test samples.

\begin{figure}[t]
\begin{center}
   \includegraphics[width=0.8\linewidth]{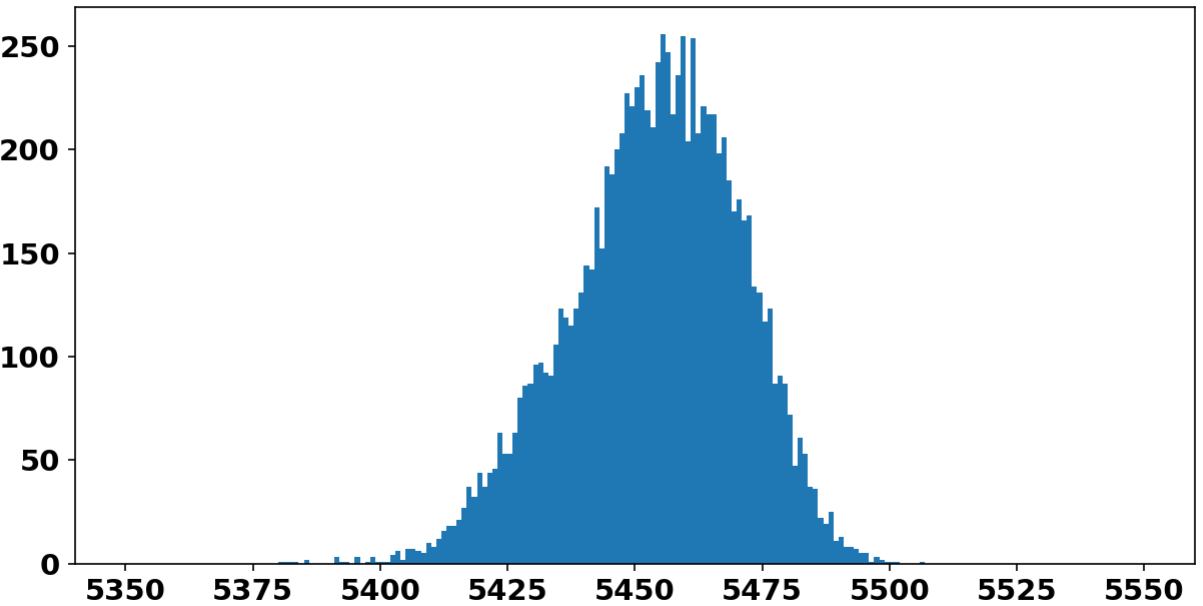}
\end{center}
\vspace{-3mm}
   \caption{Distribution: X-axis is the number of gates on, 
   while Y-axis is the number of samples.}
\label{figs.sample_vs_gate}
\vspace{-4mm}
\end{figure}

We also examined how many gates are fired when processing each test example.
The maximum and minimum number of fired gates per sample is 5380 and 5506 respectively.
The average number is around 5453. The number of gates used each time seems to obey a
normal distribution (see Figure~\ref{figs.sample_vs_gate}).

\begin{figure}[t]
\begin{center}
   \includegraphics[width=\linewidth]{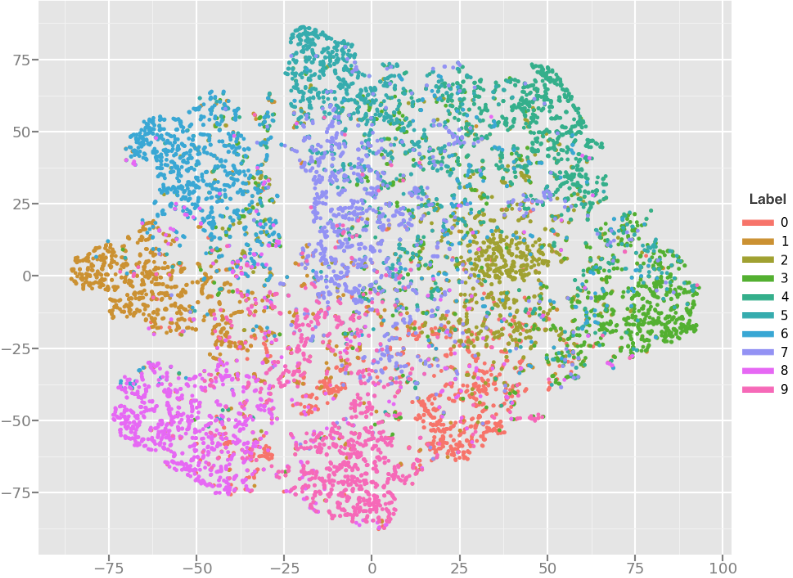}
\end{center}
\vspace{-3mm}
   \caption{Visualization of the high dimensional gate vectors. PCA and t-SNE are applied 
   to reduce the
   vector dimension to 2. Dots with the same color corresponds to test samples with the 
   same label. Better viewed in color.}
\label{figs.tsne}
\vspace{-3mm}
\end{figure}

Lastly, we want to investigate what gating strategy has been learned by the gater network. To do so, we represent the filter usage of each test sample as a 7200-dimensional binary vector where each element in the vector represents if the corresponding gate is on ($1$) or off ($0$). We collect the filter usage vector of each sample and reduce the dimension of these vectors
from 7200 to 400 using Principal Component Analysis (PCA). We then project these vectors onto a 2-dimensional space via t-SNE~\cite{maaten2008visualizing} (see Figure~\ref{figs.tsne}). 
Interestingly, we find samples of the same class tend to use similar gates. In the figure, each color of dots represents a ground-truth label. 
This shows that the gater network learned to turn on similar gates for samples from the same class---
hence similar parts of the backbone network are used to process the samples from the class. 
On the other hand, we found the clusters in Figure~\ref{figs.tsne} is still far from perfectly setting samples from different labels apart.
It is indeed a good evidence that the gater network doesn't accidentally take over the prediction task that the backbone network is intended to do, which
is what we want to avoid. We want the gater network to focus on learning to make good decisions on which 
filters in the backbone network should be used. From this analysis, we can see that the experiments are turned 
out as we expected and the backbone network still does the essential part of prediction for achieving the high 
accuracy.

We can draw the following conclusions from the above observations and analyses:
\begin{itemize}
    \vspace{-3mm}
    \item The gater network is capable of learning effective gates for different samples.
    It tends to generate similar gates for samples from the same class (label).
    \vspace{-3mm}
    \item The residual blocks at shallow layers are more redundant than those at deep layers.
    \vspace{-3mm}
    \item Input-dependent features are more needed at deep layers than at shallow layers.
\end{itemize}

\vspace{-4mm}
\subsubsection{State-of-the-Art on CIFAR-10}
\vspace{-2mm}
Next we test the performance of our method with state-of-the-art models, 
Shake-Shake~\cite{yamada2018shakedrop} and DenseNet~\cite{huang2017densely}, 
on CIFAR-10 dataset.
We use Shake-Shake and DenseNet as the backbone network and ResNet-20 again as the gater network
to form our models respectively.
Table~\ref{tab.sota_cifar} summarizes the comparison of our models with the original models.
The gater network in our method consistently improves the state-of-the-art
backbone network without significantly increasing the number of parameters. One of our 
models, {\it Shake-Shake-Gated 26 2}x{\it96d}, has only 1.2\% more parameters than the corresponding
baseline model. Another interesting finding is that, with the assistance of the gater network,
{\it DenseNet-BC-Gated} $(L=250, K=24)$ is even performing better than both 
{\it DenseNet-BC} $(L=190, K=40)$ and {\it DenseNet-BC-Gated} $(L=190, K=40)$, 
although {\it DenseNet-BC-Gated} $(L=250, K=24)$ has much fewer parameters.

Note that in~\cite{cutout}, it is shown when {\it Shake-Shake 26 2}x{\it96d} is combined with
a data pre-processing technique called {\it cutout}, it can achieve 2.56\% error rate on 
CIFAR-10 test set. The technique is orthogonal to our method and can also be combined with
our method to give better results.

\begin{table*}[t]
\caption{Classification error rates on the CIFAR-10 test set. All the methods
use data augmentation during training. All our models are using ResNet-20 as the gater network. 
The Gated Filters column shows the number of filters subject to gating in our model.
All the baseline results are from our reimplementation.}
\vspace{-3mm}
\label{tab.sota_cifar}
\begin{center}
\begin{tabular}{l|r|ll}
& \multicolumn{1}{c|}{\bf Gated Filters} & \multicolumn{1}{c}{\bf Param} & \multicolumn{1}{c}{\bf Error Rates \%} \\ \hline
{\bf DenseNet-BC} $(L=100, k=12)$~\cite{huang2017densely} &- & 0.77M & 4.48 \\
{\bf DenseNet-BC-Gated} ($L=100, k=12, {\bf Ours}$)  &540 & 1.05M & {\bf 4.03} \\ \hline 
{\bf DenseNet-BC} $(L=250, k=24)$~\cite{huang2017densely} &- & 15.32M & 3.61 \\ 
{\bf DenseNet-BC-Gated} ($L=250, k=24,{\bf Ours}$) &2,880 & 15.62M & {\bf 3.31} \\ \hline 
{\bf DenseNet-BC} $(L=190, k=40)$~\cite{huang2017densely} &- & 25.62M & 3.52 \\ 
{\bf DenseNet-BC-Gated} ($L=190, k=40, {\bf Ours}$) &3,600 & 25.93M & {\bf 3.39}\\ \hline 
{\bf Shake-Shake} 26 2x64d~\cite{yamada2018shakedrop} &- & 11.71M & 3.05 \\
{\bf Shake-Shake-Gated} 26 2x64d ({\bf Ours}) &3,584 & 12.01M & {\bf 2.89} \\ \hline 
{\bf Shake-Shake} 26 2x96d~\cite{yamada2018shakedrop} &- & 26.33M & 2.82 \\ 
{\bf Shake-Shake-Gated} 26 2x96d ({\bf Ours}) &5,376 & 26.65M & {\bf 2.64}
\end{tabular}
\end{center}
\vspace{-2mm}
\end{table*}

\subsection{ImageNet}
\paragraph{Classification Results}
To test the performance of our method on large dataset, we apply our method to models for
ImageNet 2012. We use ResNet~\cite{he2016deep} and Inception-v4~\cite{Szegedy2017Inceptionv4IA}  
as the backbone network and ResNet-18~\cite{he2016deep} as the gater network to form our models.
Table~\ref{tab.err-imagenet} shows the classification results on ImageNet validation set 
with baselines similar
to the settings in Table~\ref{tab.resnet_cifar}.
We can see that, our method improves all the models by 0.52\% $\sim$ 1.85\% in terms of top-1 
error rate, and 0.14\% $\sim$ 0.78\% in terms of top-5 error rate. 
Note that~\cite{veitadaptive} proposes to dynamically skip layers in ResNet-101, and the 
top-1 and top-5 error rates of their model are 22.63\% and 6.26\% respectively. 
Our ResNet-101-Gated achives 21.51\% and 5.72\% on the same task, which is apparently much better than 
their model.
In addition, there are also two interesting findings:
\begin{itemize}
    \vspace{-2mm}
    \item The performance of ResNet-101 is significantly boosted with the help of the 
    gater network. ResNet-101-Gated is even performing better than ResNet-152 using much 
    fewer layers.
    \vspace{-2mm}
    \item Similar to the results on CIFAR datasets, ResNet-Wider is performing well when the
    original model is shallow and small, but is outperformed by our models when the original 
    model contains enough filters.
    \vspace{-3mm}
\end{itemize}

\begin{table*}[h] 
\caption{Single-crop error rates on the ImageNet 2012 validation set. ResNet-Wider is 
the ResNet with additional filters at each layer such that it contains roughly 
the same number of parameters as our model. ResNet-SE is the ResNet with 
squeeze-and-excitation units. All the ResNet-Gated models are using ResNet-18 as the 
gater network. The Gated Filters column shows the number of filters subject to gating 
in our model.
All the baseline results are from our reimplementation, except $^{\dagger}$ from 
the original paper.}
\vspace{-2mm}
\label{tab.err-imagenet}
\begin{center}
\begin{tabular}{l|r|lll}
& \multicolumn{1}{c|}{\bf Gated Filters} & \multicolumn{1}{c}{\bf Parameters}  &\multicolumn{1}{c}{\bf Top-1 Error \%} & \multicolumn{1}{c}{\bf Top-5 Error \%} \\ \hline
{\bf ResNet-34}~\cite{he2016deep}           &- & 21.80M & 26.56 & 8.48  \\
{\bf ResNet-34-Wider}                       &- & 33.89M & {\bf 25.36} & {\bf 7.91} \\
{\bf ResNet-34-SE}~\cite{hu2018senet}       &- & 21.96M & 26.08 & 8.30 \\
{\bf ResNet-34-Gated (Ours)}                &3,776  & 34.08M & 26.04 ($\downarrow$0.52) & 8.34 ($\downarrow$0.14) \\ \hline
{\bf ResNet-101}~\cite{he2016deep}          &- & 44.55M & 23.36 & 6.56  \\
{\bf ResNet-101-Wider}                      &- & 59.17M & 21.89 & 6.05  \\
{\bf ResNet-101-SE}~\cite{hu2018senet}      &- & 49.33M & 22.38$^\dagger$ & 6.07$^\dagger$ \\
{\bf ResNet-101-Gated (Ours)}               &32,512 & 64.21M & {\bf 21.51} ($\downarrow$1.85) & {\bf 5.78} ($\downarrow$0.78)  \\ \hline
{\bf ResNet-152}~\cite{he2016deep}          &- & 60.19M & 22.34 & 6.22  \\
{\bf ResNet-152-Wider}                      &- & 81.37M & 21.50 & 5.67  \\
{\bf ResNet-152-SE}~\cite{hu2018senet}      &- & 66.82M & 21.57$^\dagger$ & 5.73$^\dagger$ \\
{\bf ResNet-152-Gated (Ours)}               &47,872  & 83.80M & {\bf 21.19} ($\downarrow$1.15) & {\bf 5.45} ($\downarrow$0.77)    \\ \hline
{\bf Inception-v4}~\cite{Szegedy2017Inceptionv4IA}  & - &44.50M  & 20.33 & 4.99 \\ 
{\bf Inception-v4-Gated (Ours)}                        & 16,608 &61.67M & {\bf 19.64} ($\downarrow$0.69) & {\bf 4.80} ($\downarrow$0.19) %\\ \hline
%{\bf NasNet-A}~\cite{Oquab2014LearningAT} \\ 
%{\bf NasNet-A-Gated (Ours)}
\end{tabular}
\end{center}
\vspace{-5mm}
\end{table*}

\subsection{Implementation Details}
We train the baseline model for each architecture by following the training scheme proposed in the original papers. We pre-train the backbone and the gater network on the target task separately to properly initialize the weights. 
The training scheme here includes training configurations such as the number of training epochs, 
the learning rate, the batch size, the weight decay and so on.

After pre-training, we train the backbone and the gater network jointly as a single model. In addition to following the original training scheme for each
backbone architecture, we introduce a few minor modifications. Firstly, we increase the number of training epochs for DenseNet-Gated and Shake-Shake-Gated
by 20 and 30 respectively as they seem to converge slowly at the end of training. Secondly, we set the initial learning rate for DenseNet-Gated and Shake-Shake-Gated to a smaller value, 0.05, since a large learning rate seem to result in bad performance at the end.

Note that not all the filters in a backbone network are subject to gating in our experiments. When ResNet is used as the backbone, we apply filter selection to the last convolutional layer in each residual unit, which is similar to the SE-block in~\cite{hu2018senet}. As for DenseNet, we apply filter selection to all the convolutional layers except the first in each dense block. There are multiple residual branches in each residual block in Shake-Shake. We apply filter selection to the last convolutional layer in each branch of the residual block in Shake-Shake. In Inception-v4, there are many channel concatenation operations. We apply filter selection to all the feature maps after the channel concatenation operations.

For all our models on CIFAR, we set the size of the bottleneck layer to 8.
ResNet-34-Gated, ResNet-101-Gated and Inception-v4-Gated use a bottlneck size of 256, while ResNet-152-Gated uses 1024.

\section{Conclusions}
In this paper, we have proposed GaterNet, a novel architecture for input-dependent dynamic filter selection in CNNs. It involves two distinct components: a backbone network that conducts actual prediction and a gater network that decides which part of backbone network should be used for processing each input. Extensive experiments on CIFAR and ImageNet indicate that our models consistently outperform the original models with a large margin. On CIFAR-10, our model improves upon state-of-the-art results. We have also performed an in-depth analysis about the model behavior that reveals an intuitive gating strategy learned by the gater network.

\section{Acknowledgments} We want to thank Sergey Ioffe, Noam Shazeer, Zhenhai Zhu and anonymous reviewers for their insightful feedback.

{\small
\bibliographystyle{ieee_fullname}
\bibliography{arxiv}
}

\newpage
\appendix
\section*{Appendix}
\section{Distribution of Gates}
The amounts of filters in different residual units of ResNet-164 are different.
To give a complete view of the gates that our gater is learning, 
we plot the same figure as Figure~\ref{figs.gate_dist}
with the number of gates on the Y-axis in Figure~\ref{figs.gate_dist_count} below.
A gate is an entry in the binary gate vector $g$. It
corresponds to a filter in the backbone network ResNet-164.
A gate is {\it always off} means that it is 0 for all the samples in the
test set. A gate is {\it always on} means that it is 1 for all the test
samples. 
And a gate
is {\it input-dependent} means that it is 1 for some of the test samples
and 0 for the others.

\begin{figure}[h]
\begin{center}
  \includegraphics[width=\linewidth]{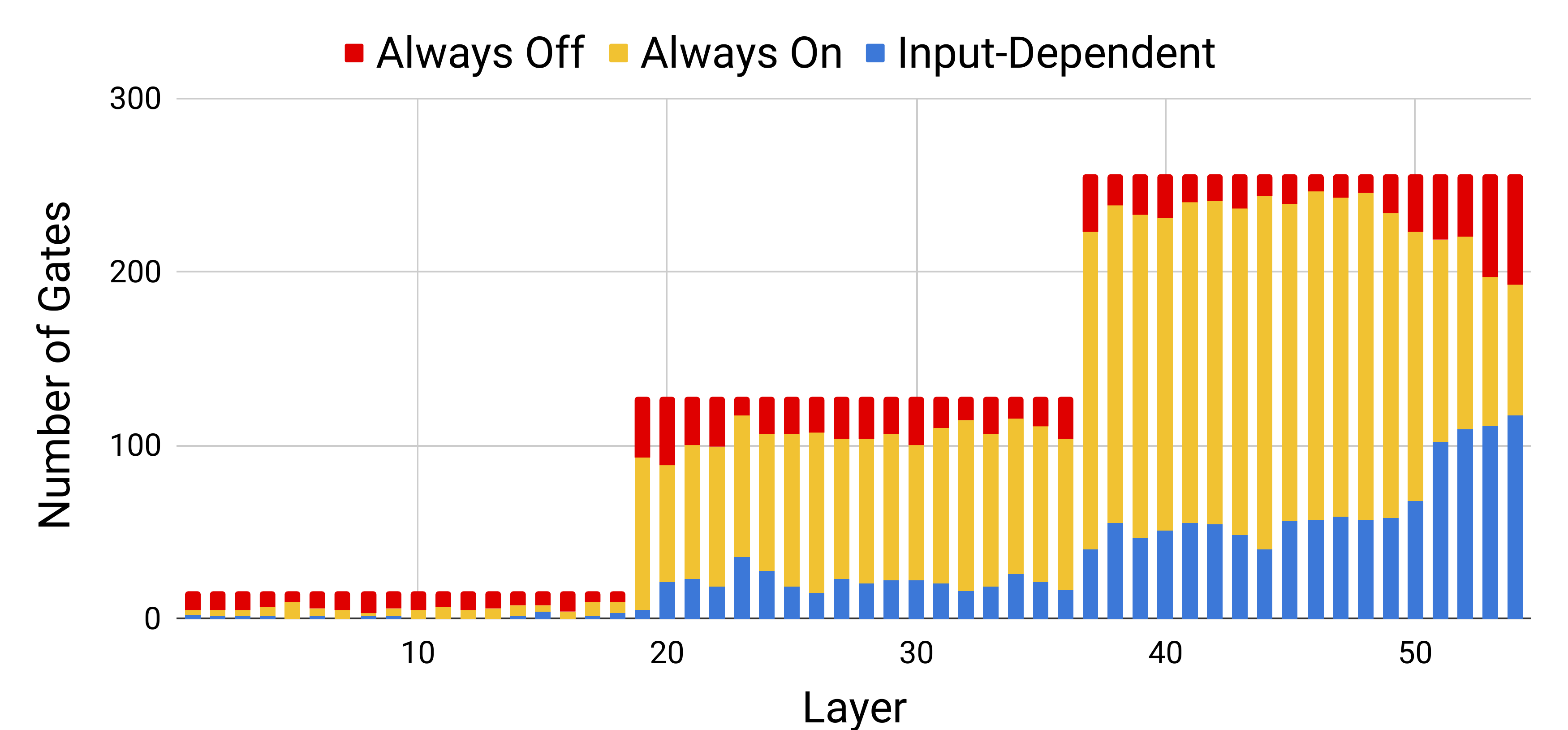}
\end{center}
  \caption{Number of the three type of gates for different layers
  in the backbone ResNet-164 of ResNet-164-Gated on Cifar-10 Test Set. There
  are totally 54 residual units in ResNet-164. Better viewed in color.}
\label{figs.gate_dist_count}
\end{figure}

\section{Scheduled Dropout}
In the experiments of DenseNet-Gated, Shake-Shake-Gated and Inception-v4-Gated,
scheduled dropout~\cite{JMLR:v15:srivastava14a} similar to ScheduledDropPath 
in~\cite{Zoph_2018_CVPR} are applied to the gate vector $g$.
We start from a dropout rate of 0.0 and increase it gradually during training.
The dropout rate reaches 0.05 at the end of training.

\end{document}